\documentclass{bmvc2k}

\usepackage{amssymb}
\usepackage{multirow}
\usepackage{booktabs}

\title{Uncovering Hidden Challenges in Query-Based Video Moment Retrieval}

\addauthor{Mayu Otani}{otani_mayu@cyberagent.co.jp}{1}
\addauthor{Yuta Nakashima}{n-yuta@ids.osaka-u.ac.jp}{2}
\addauthor{Esa Rahtu}{esa.rahtu@tuni.fi}{3}
\addauthor{Janne Heikkil\"{a}}{janne.heikkila@oulu.fi}{4}

\addinstitution{
 CyberAgent, Inc.\\
 Tokyo, Japan
}
\addinstitution{
 Osaka University\\
 Osaka, Japan
}
\addinstitution{
 Tampere University\\
 Tampere, Finland
}
\addinstitution{
 University of Oulu\\
 Oulu, Finland
}

\runninghead{Otani, Nakashima, Rahtu, Heikkil\"{a}}{Challenges in Video Moment Retrieval}

\def\etal{\emph{et al}\bmvaOneDot}
\def\ie{\emph{i.e}\bmvaOneDot}

\begin{document}

\maketitle

\begin{abstract}
The query-based moment retrieval is a problem of localising a specific clip from an untrimmed video according a query sentence. This is a challenging task that requires interpretation of both the natural language query and the video content. Like in many other areas in computer vision and machine learning, the progress in query-based moment retrieval is heavily driven by the benchmark datasets and, therefore, their quality has significant impact on the field. In this paper, we present a series of experiments assessing how well the benchmark results reflect the true progress in solving the moment retrieval task. Our results indicate substantial biases in the popular datasets and unexpected behaviour of the state-of-the-art models. Moreover, we present new sanity check experiments and approaches for visualising the results. Finally, we suggest possible directions to improve the moment retrieval in the future. Our code for this paper is available at \url{https://mayu-ot.github.io/hidden-challenges-MR/}.
\end{abstract}

\section{Introduction}
The capability of retrieving specific events from video content is an appealing property for many practical applications. However, the underlying search problem is very challenging due to the complicated nature of the possible activities and queries. For this reason, the approaches relying on predefined object or action classes are not well suited for this problem. Therefore, a relatively new research area called query-based moment retrieval has gained plenty of interest \cite{Gao_2017_ICCV,hendricks17iccv,10.1145/3209978.3210003,yuan2019semantic,2DTAN_2020_AAAI} in the computer vision community. 

\begin{figure}[t!]
    \centering
    \includegraphics[width=0.9\linewidth, clip]{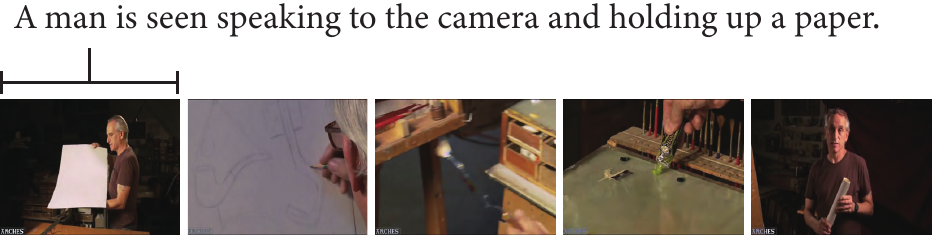}
    \caption{Moment retrieval finds the moment in a video corresponding to a query sentence.}
    \label{fig:example}
\end{figure}

The query-based moment retrieval in videos, also known as temporal sentence grounding or moment retrieval, aims at locating a specific moment from the input sequence that matches to the given query sentence (see Fig.~\ref{fig:example}). Similarly to many other fields in machine learning, the development in the query-based moment retrieval is heavily driven by the vision and language benchmark datasets such as TACoS \cite{regneri-etal-2013-grounding}, DiDeMo \cite{hendricks17iccv}, Charades-STA \cite{Gao_2017_ICCV}, and ActivityNet Captions \cite{krishna2017dense}. In particular, the latter two have been widely adopted as the standard benchmarks in the recent works \cite{10.1145/3209978.3210003,Gao_2017_ICCV,DBLP:conf/aaai/YuanM019,DBLP:conf/aaai/Xu0PSSS19,Hahn2019,2DTAN_2020_AAAI,yuan2019semantic}.

It is vital that the benchmark results reflect the true progress in solving the original problem, otherwise the entire field the can be steered into a wrong direction. Despite the importance, this aspect is not well studied, particularly, in the context of query-based moment retrieval. Inspired by the similar works in other domains \cite{Dawson2018,balanced_vqa_v2}, we perform in depth analysis on the recent models and benchmark datasets in moment retrieval. In particular, we use Charades-STA \cite{Gao_2017_ICCV} and ActivityNet Captions \cite{krishna2017dense} datasets for our study. %

The main findings of this work include:

\begin{itemize}
\item \textbf{Popular datasets include significant biases.} We observe that the query sentences provide a strong prior on the temporal locations of the moments. Based on this observation, we develop a set of simple baseline models which do not use any visual content and show that they obtain non-trivial performance surpassing numerous recent works. 

\item \textbf{State-of-the-art models are often agnostic to video.} We find evidence that the current state-of-the-art models do not necessarily make any (or very little) use of the visual input. In particular, with ActivityNet Captions, the models do not learn cross-modal matching, but exploit the dataset biases instead. 

\item \textbf{Limitations in the current benchmarks.} We investigate human performance on the same tasks and find it lower than with the state-of-the-art models. Moreover, we discover substantial disagreement between different annotators, which may indicate that the visual task proposed in the datasets is highly ambiguous. 
\end{itemize}

The evaluation code and the additional annotations will be made publicly available. We hope that our work inspires similar analysis for other vision and language tasks.

\section{Related work}

\paragraph{Dataset analysis} This paper is inspired by works that analyze benchmarks for visual understanding tasks, such as VQA \cite{balanced_vqa_v2,agrawal-etal-2016-analyzing} and action recognition/detection \cite{Sigurdsson2017, Alwassel2018}.
For VQA, biases in a popular dataset and models' capability of image understanding are investigated in \cite{balanced_vqa_v2,agrawal-etal-2016-analyzing}.
The Charades dataset \cite{Sigurdsson2016}, upon which Charades-STA is built, is also investigated in \cite{Sigurdsson2017} with respect to the action recognition task.
Our work also analyzes benchmarks for video moment retrieval from the aspect of dataset biases and check if models make use of input video.

Some prior works for moment retrieval have already pointed out that there are biases in datasets \cite{yuan2019semantic,1907.12763}.
Yuan \etal reported that randomly selecting candidate moments performs well due to biases on Charades-STA \cite{yuan2019semantic}. Escorcia \etal indicates strong biases between temporal locations and query sentences in DiDeMo and Charades-STA \cite{1907.12763}. They demonstrate simply incorporating the temporal location of a moment significantly boosts the performance. Furthermore, they also provide a strong baseline using only temporal locations and query sentences, which is closely related to our blind baselines.
Our analysis further adds new insights about the dataset by revealing the biases in temporal annotation and how the biases affect the moment retrieval performance.

\paragraph{Video moment retrieval} A two-stage framework has been used for the moment retrieval task that generates candidate video moments and ranks them according to the relevance to the query sentence.
Some methods extract candidate moments with temporal sliding windows and predict the relevance \cite{Gao_2017_ICCV,10.1145/3209978.3210003}. They also refine the moments' boundaries by predicting the offsets to ground truth boundaries.
Some other methods use a model to generate candidate moment boundaries \cite{DBLP:conf/aaai/Xu0PSSS19}.
Inspired by recent object and action detection methods, a single-shot framework has also been explored \cite{2DTAN_2020_AAAI,yuan2019semantic}.
This framework skips candidate moment generation and encodes video moments in different durations in a single pass.
Alternatively, Yuan \etal \cite{DBLP:conf/aaai/YuanM019} proposed to regress temporal locations of a moment from the query sentence and the global feature of the target video.
Hahn \etal \cite{Hahn2019} trained an agent that adjusts moment boundaries in the reinforcement learning framework.
This paper presents an analysis of the performance of these methods by comparing them to our baseline models.
We provide further investigation about the behavior of the methods \cite{2DTAN_2020_AAAI,yuan2019semantic}.

\section{Dataset analysis}
\label{sec:dataset_analysis}
\begin{figure}[t!]
\begin{tabular}{c}
    \centering
    \begin{minipage}{0.49\linewidth}
    \centering
    Charades-STA \\
    \includegraphics[width=\linewidth, clip]{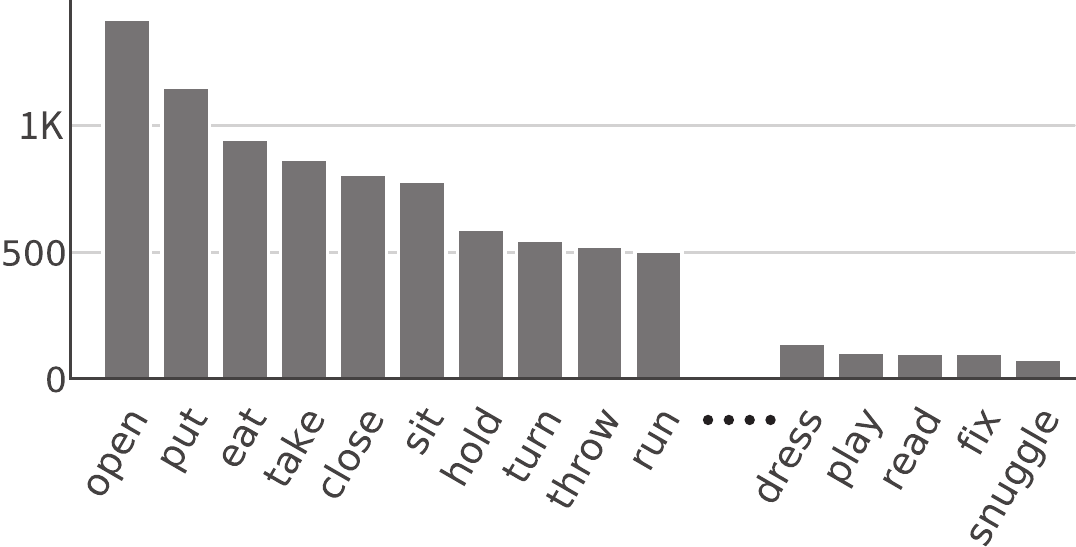}
    \end{minipage}
    \begin{minipage}{0.49\linewidth}
    \centering
    ActivityNet Captions \\
    \includegraphics[width=\linewidth, clip]{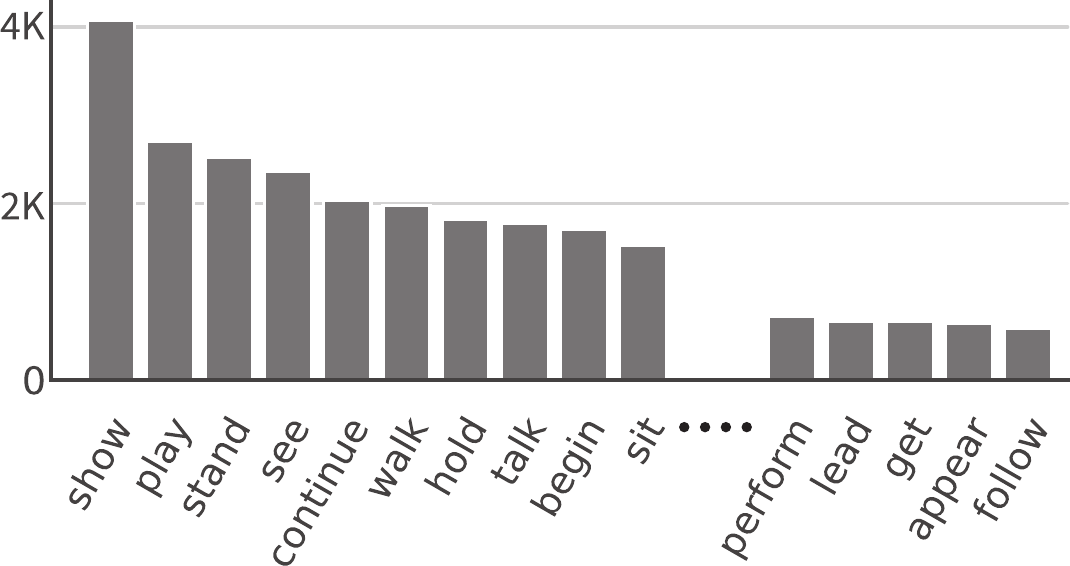}
    \end{minipage}
\end{tabular}
    \caption{Top-30 frequent actions of each dataset.}
    \label{fig:action_vocab}
\end{figure}

We perform our analysis using Charades-STA \cite{Gao_2017_ICCV} and ActivityNet Captions \cite{krishna2017dense} datasets.

\textbf{Charades-STA} is built upon Charades \cite{Sigurdsson2016} and contains 9,848 videos, each of which is associated with multiple natural language sentences.
Each sentence has a temporal annotation that indicate the start and end points of the corresponding moment in the video.
The videos are created by asking crowd sourcing workers to record themselves performing actions based on a short script (\ie, a set of the sentences associated with the video), where the script was written by composing predefined vocabulary and describe multiple daily actions. 

\textbf{ActivityNet Captions} contains 19,209 YouTube videos. Each video is associated with captions and their temporal locations. This dataset was originally tailored for dense video captioning, but has been recently used also for moment retrieval. Currently, this is the largest dataset for the video moment retrieval task. Typical query sentences are longer than Charades-STA's and often describe multiple actions as shown in Fig.~\ref{fig:example}.

\subsection{Biases in query sentences and moment locations}
We start by analyzing the dataset biases by exploring the query sentences. %
The query sentence often describe an action of a person. Therefore, we extract the verbs from query sentences to provide an overview of what actions are in the datasets.
Figure~\ref{fig:action_vocab} shows the top-30 frequent verbs, which cover 93.2\% of all verbs (or actions) in Charades-STA and 51.4\% in ActivityNet Captions.
For Charades-STA, the annotators were asked to write sentences using predefined vocabulary, and so the diversity of sentences is rather limited.
ActivityNet Captions uses the verb ``show'' frequently, but this word is less likely to describe actions of person as there are many sentences like ``something is \textit{shown} to the camera.''
We can observe that some verbs are used more frequently than others, which implies that they might have larger impact on evaluation scores.

\begin{figure}[t]
    \centering
    Charades-STA \\
    \includegraphics[clip,width=0.9\linewidth]{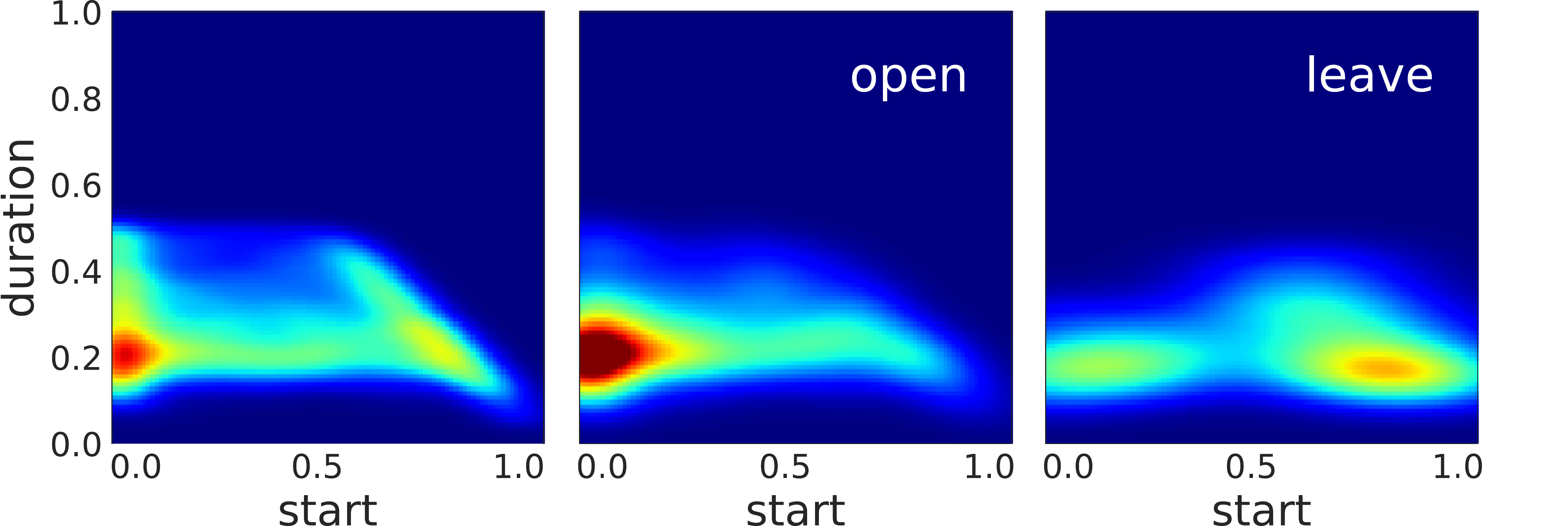} \\

    ActivityNet Captions\\
    \includegraphics[clip,width=0.9\linewidth]{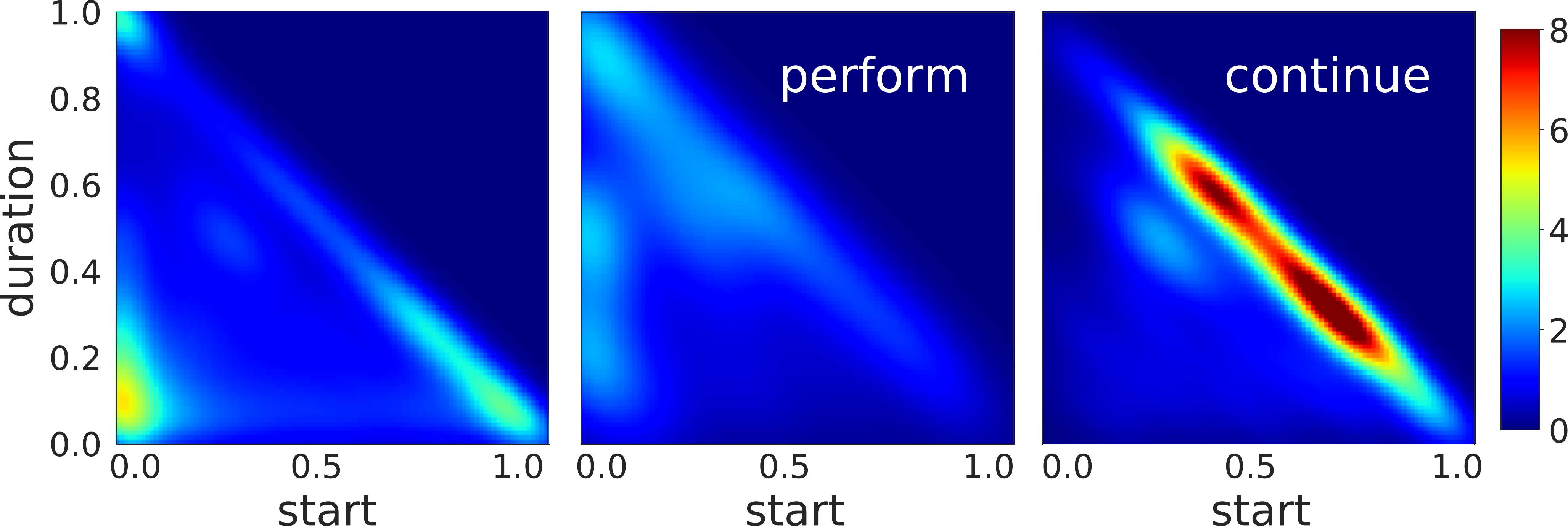}
    \caption{Distributions of temporal locations of target moments. Color represents values of probability density function. The top three plots are the distributions for Charades-STA, and the bottom three are for ActivityNet Captions. For each dataset, the left distribution is produced for all moments, while the other two distributions are moments described by a certain verbs. More examples can be found in the supplementary material.}
    \label{fig:moment_location}
\end{figure}

One important characteristics of these datasets lies in the biases of temporal locations of target moments.
The left density plots in Fig.~\ref{fig:moment_location} show the overall distributions of the temporal moment locations in Charades-STA and ActivityNet Captions, where the horizontal and vertical axes are the starting time and duration of a moment, both of which are normalized to the range of 0 to 1 by dividing them by the length of the respective video.
These distributions are obtained using kernel density estimation with the Gaussian kernel. In Charades-STA, target moments are more likely to start at the very beginning of videos and last roughly 20\% of the video length.
For ActivityNet Captions, we can see a peak near the bottom left corner of the plot.
This peak corresponds to moments that start at the beginning of videos and cover roughly 10\% of the video length.
These biases provide powerful priors on moment locations.
Weighting candidate locations according to these distributions can improve the chances of obtaining the correct moment.

We hypothesize that verbs in query sentences provide further hints on the temporal locations of target moments.
For example, a moment described by ``cooking something'' may be longer than one described by ``throwing something.''
We thus estimated the distribution of temporal locations for each verb.
Examples are also shown in Fig.~\ref{fig:moment_location}.
In Charades-STA, target moments corresponding to``open'' are biased to start at the beginning of videos, while ones corresponding to ``leave'' are often located at the end.
ActivityNet Captions also shows biases.
These results suggest that we can guess the temporal location of target moments more precisely by using only one word (\ie, the verb) in a query sentence.

\subsection{Analysis with blind baselines}

Given the observation above, the question is how far one can go using only the priors (\ie, without using any visual information). To this end, we implement three blind baseline models that exploit the biases at different levels. 

\paragraph{Prior-Only Blind} The first baseline predicts temporal locations without using videos or query sentences. %
This baseline randomly samples 100 temporal locations from the prior distribution, which is computed from all training samples (the top-left and bottom-left distributions in Fig.~\ref{fig:moment_location}).
The sampled locations are ranked based on their likelihood, and the most probable sample is chosen.\footnote{This baseline's first-ranked location approaches to the starting time and duration at the mode of the prior distribution when the number of sample increases.}

\paragraph{Action-Aware Blind} This baseline uses only one word in a query sentence to predict temporal locations of the moments. As query sentences often describe humans' actions, we exploit keywords corresponding to actions. For simplicity, we use the first verb in a query sentence as an approximation of the target action. Although these verbs do not always correspond to the true action, the main purpose is to provide a simple yet powerful clue for predicting the temporal locations of the moments exploiting the biases. The distribution of locations given the verb in the query sentences is computed beforehand for the top-50 frequent verbs. At inference time, this baseline takes the first verb from the query sentence and samples 100 temporal locations from the corresponding conditional distribution.
For verbs that are not in top-50, we use the Prior-Only Blind baseline, because the corresponding conditional distributions might be unreliable.

\paragraph{Blind-TAN} We implement a neural network-based model that uses the full query sentence to predict temporal locations. Blind-TAN is built upon 2D-TAN~\cite{2DTAN_2020_AAAI}, which encodes a query sentence with an LSTM network. Video features of candidate moments (\ie, all possible temporal segments formed by uniformly sampled starting time and end time) are extracted with a pre-trained CNN, forming a two-dimensional map of visual features (the first and second dimensions of the map correspond to starting and end times). The sentence and visual features are fused and  fed into a CNN to produces a map, each value of which is the score of the corresponding candidate moment.
Blind-TAN removes the CNN to extract video features and replaces the map of visual features with a learnable map in the same shape. By training this model solely with query sentences, the learnable map may acquire some ideas on when certain actions are likely to happen.
More details are provided in the supplementary material. \\

\noindent The evaluation is performed using popular $\mathrm{R}@k(\mathrm{IoU}>m)$, which is the percentage of query sentences in the test set that have at least one moment in top-$k$ retrieved moments with IoU larger than $m$. Following the recent works, we report R@1(IoU$>$0.5).
\begin{figure}[t]
    \centering
    \begin{tabular}{c}
    
    \begin{minipage}{0.49\hsize}
    \centering
    Charades-STA \\
    \includegraphics[width=\linewidth,clip]{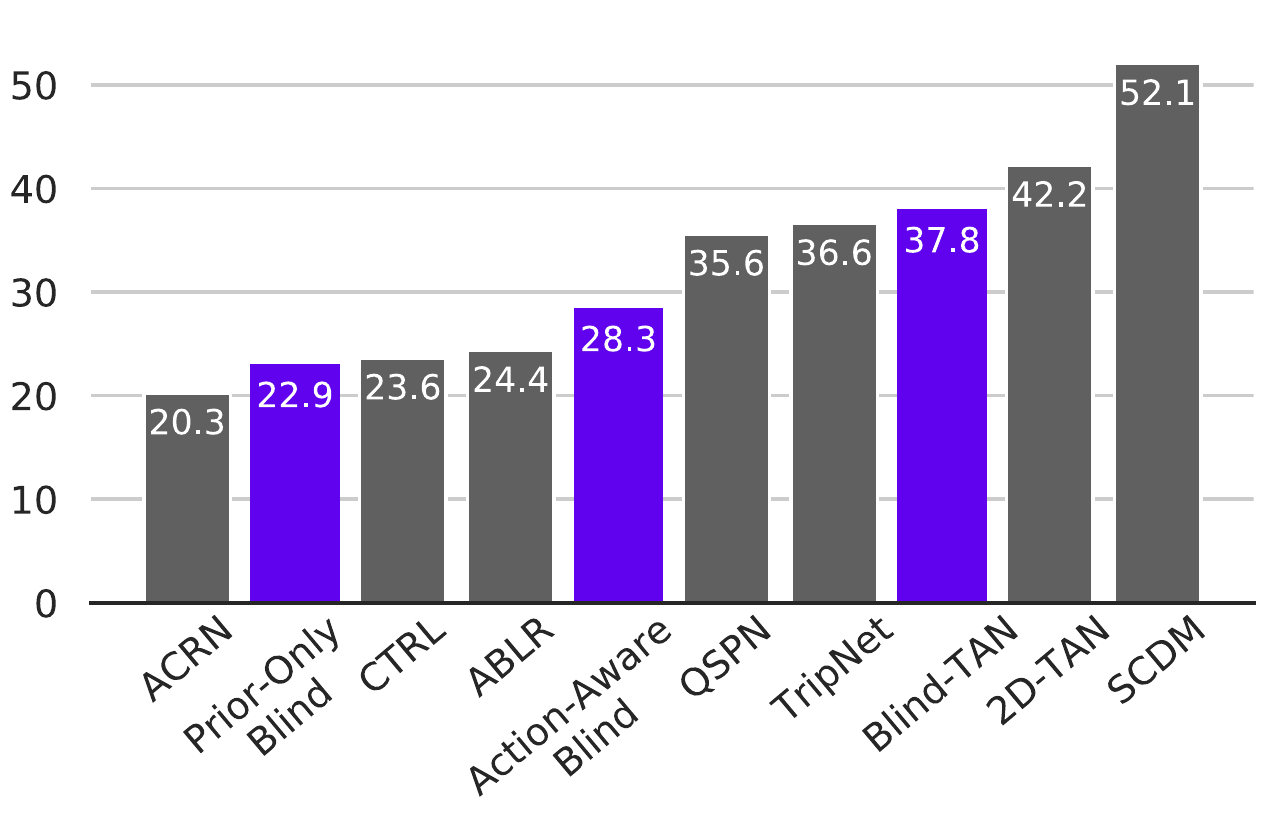}
    \end{minipage}
    
    \begin{minipage}{0.49\hsize}
    \centering
    ActivityNet Captions \\
    \includegraphics[width=\linewidth,clip]{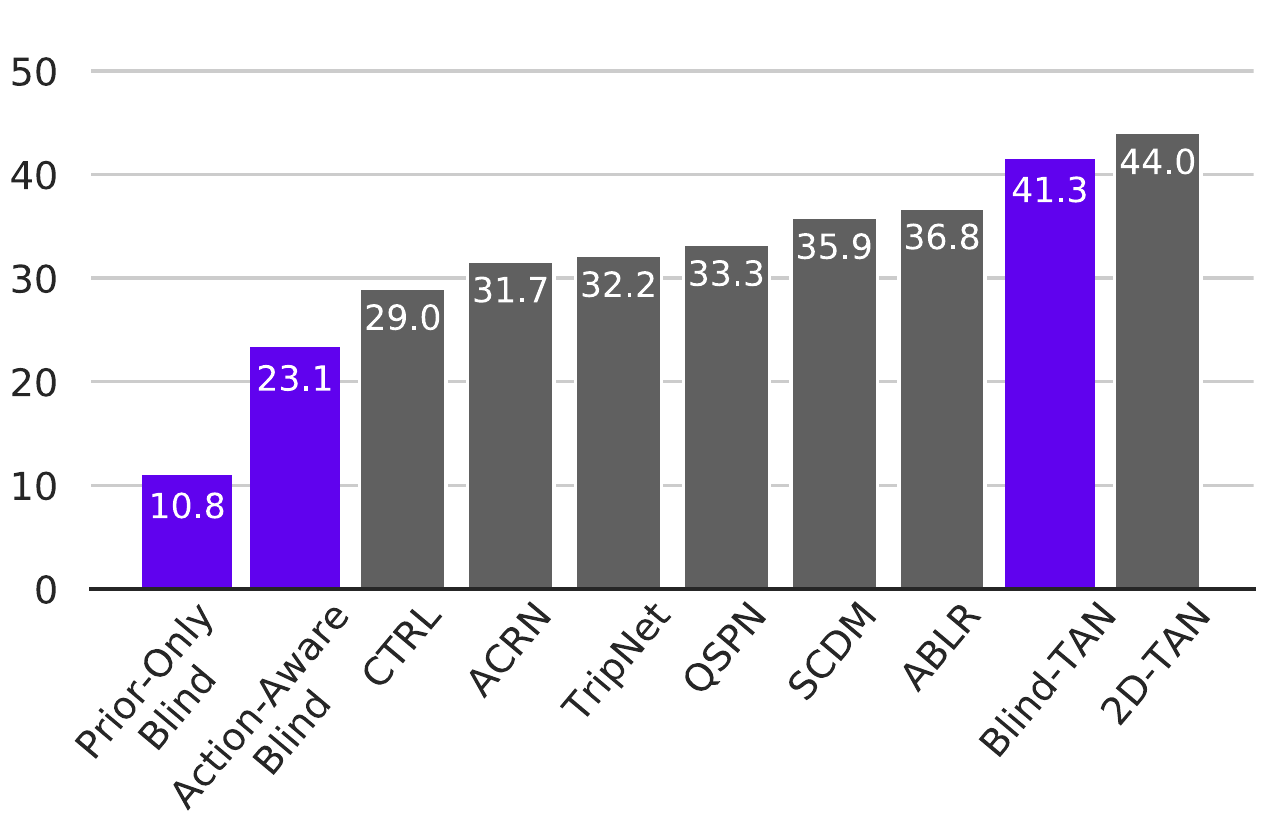}
    \end{minipage}
    
    \end{tabular}
    
    \caption{R@1 (IoU$>$0.5) scores on Charades-STA (left) and ActivityNet Captions (right). Highlighted bars indicate blind baselines. Surprisingly, the blind baselines outperform many deep models and reach close to the state-of-the-art on ActivityNet Captions.}
    \label{fig:baseline_performance}
\end{figure}
Figure~\ref{fig:baseline_performance} summarizes the scores for our blind baselines as well as for the recent models, namely
ACRN \cite{10.1145/3209978.3210003}, 
CTRL \cite{Gao_2017_ICCV}, 
ABLR \cite{DBLP:conf/aaai/YuanM019}, 
QSPN \cite{DBLP:conf/aaai/Xu0PSSS19}, 
TripNet \cite{Hahn2019}, 
2D-TAN \cite{2DTAN_2020_AAAI}, and 
SCDM \cite{yuan2019semantic}.
The scores for these models are taken from the original papers, except for SCDM. The SCDM's evaluation discards some samples and thus the reported score is not comparable. Therefore, we recomputed the scores using all test samples. In addition, we tested an option which samples temporal locations from the uniform distribution in $[0, 1]$.\footnote{This may generate invalid combinations of starting and end times. We keep sampling a temporal location until we have valid one.} The scores of the baseline without priors is computed by computing the score for entire test set 100 times and averaging them. As a result, we obtain 10.77\% on Charades-STA and 13.57\% on ActivityNet. 

Surprisingly, our blind baselines are competitive and even outperform some deep models on Charades-STA.
The action-aware baseline outperforms ACRN, CTRL, ABLR, and other recent works  \cite{DBLP:conf/aaai/ChenJ19a,10.1145/3323873.3325019,Wang_2019_CVPR} that were not included in the figure due to space constraints.
On ActivityNet Captions, Prior-Only Blind and Action-Aware Blind do not perform as well as the deep models, but Blind-TAN achieves a score close to the state-of-the-art.
This may be because the query sentence in ActivityNet Captions often describes some actions (2.2 verbs appear on average), whereas our Action-Aware Blind used only the first verb from the query. Using more information from the query led to further improvements to the baseline scores.

In summary, the baseline results indicate the significance of the biases in the datasets, which should be considered when assessing the models.
Many recent works compare their models to other deep models; however, these comparisons might not be that meaningful if the comparison model performs below the blind baseline. 
This result implies that building strong baselines is essential in order to validate deep models in general.

Furthermore, some deep models outperform our baselines, but unfortunately, this does not necessarily mean that the deep models are actually predicting based on video. We will investigate this further in the following section.

\section{Sanity check on visual input}
\label{sec:sanity_check}

\begin{figure}[t!]
\centering
\includegraphics[clip,width=0.8\linewidth]{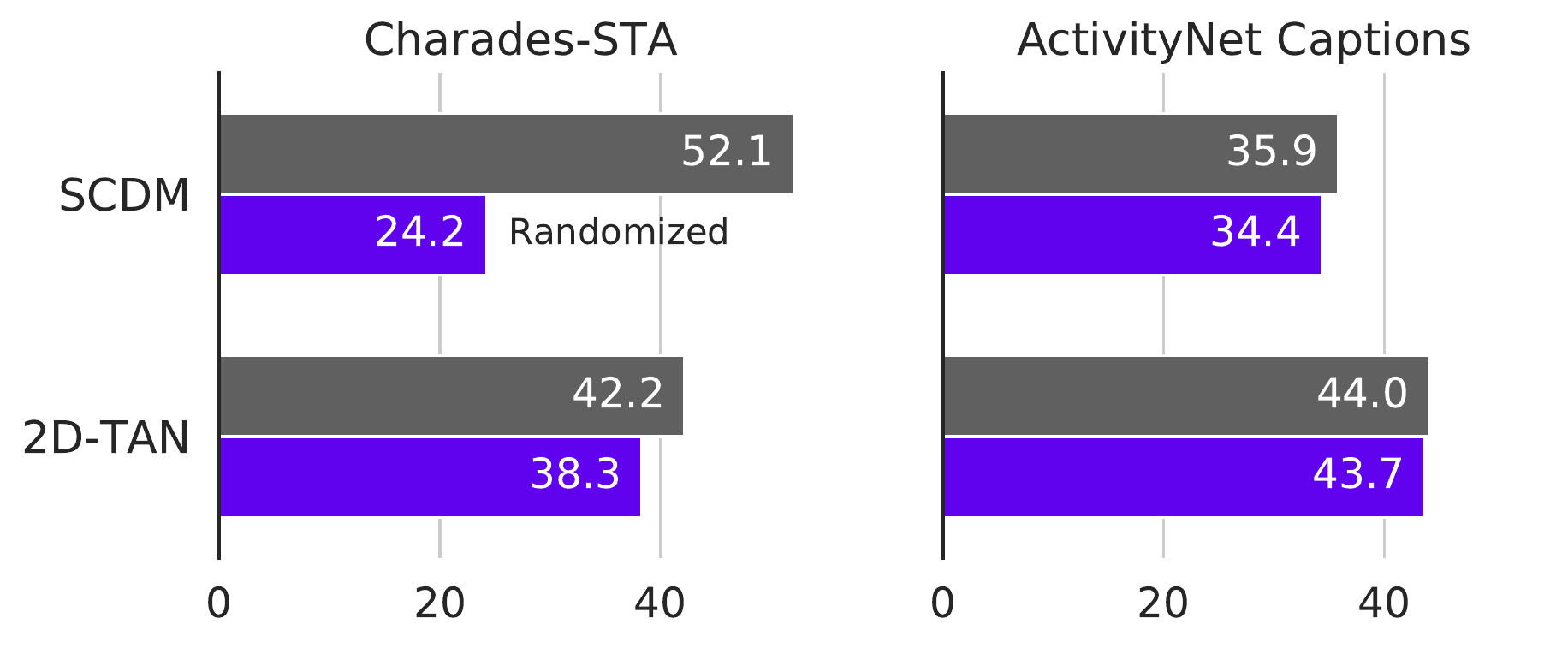}
\caption{R@1(IoU$>$0.5) scores for 2D-TAN \cite{2DTAN_2020_AAAI} and 
SCDM \cite{yuan2019semantic} when the original input videos and randomized ones are fed into these models. }
\label{fig:sanity_check}
\end{figure}

\begin{figure}[t!]
    \centering
    \includegraphics[width=0.95\linewidth,clip]{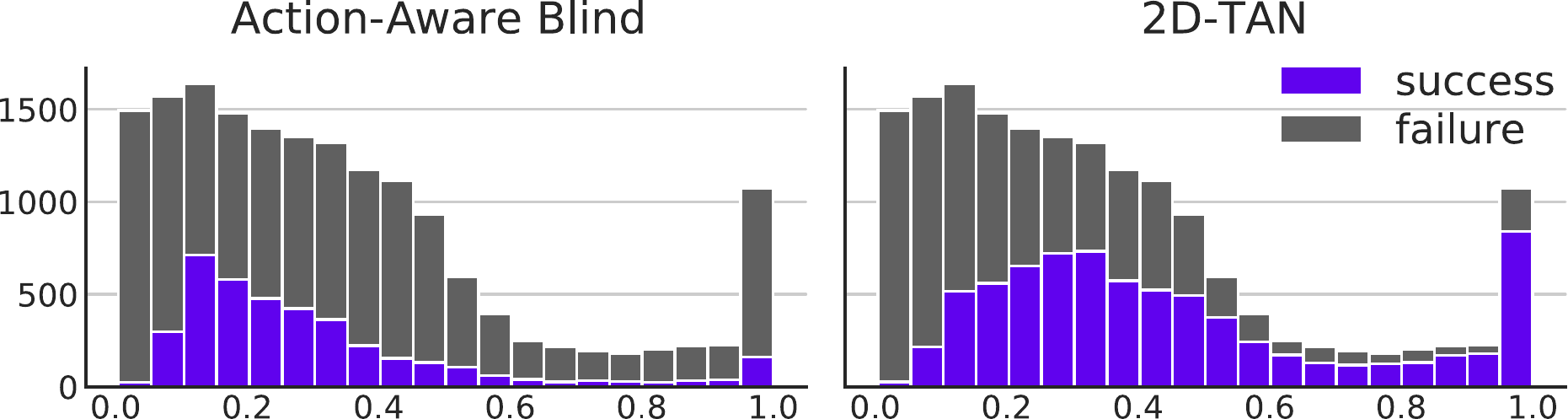}
    \caption{The number of success and failure cases with respect to the duration of the ground-truth moment on ActivityNet Captions.}
    \label{fig:performance_per_duration}
\end{figure}

As observed in the previous section (Fig.~\ref{fig:baseline_performance}), many deep models do not perform better than our blind baselines. 
This suggests that the deep models might also heavily rely on the priors, and the contribution of visual input may not be significant.
To clarify this, we experimentally show how much recent deep models take input videos into account for prediction.

To this end, we divide input videos into short segments and randomly reorder them before evaluating the models (note that we used the original input videos for training).
This randomization messes up the correspondence between input videos and ground truth temporal locations of target moments.
If a model's prediction  is based on input videos, the performance should drop significantly by this randomization;
otherwise, we can conclude that input videos do not help prediction.

We tested two recent models: SCDM \cite{yuan2019semantic} and 2D-TAN \cite{2DTAN_2020_AAAI}, which achieve the state-of-the-art performances on Charades-STA and ActivityNet Captions, respectively. 
For 2D-TAN, we used a model with trained parameters provided by the authors. For SCDM, we used authors' implementation but trained a model. We confirmed that we could reproduce the performance close to the original reported score. 

Figure \ref{fig:sanity_check} shows the scores when the original input videos and randomized ones are used.\footnote{The scores for the original input videos are adopted from the corresponding papers.} 
SCDM shows a significant performance drop for randomized videos on Charades-STA, which implies that SCDM uses visual information for prediction. Interestingly, however, SCDM's score for randomized input videos are on a par with the original input video on ActivityNet Captions.
To see the effect of input videos on prediction, we computed differences between the starting (or end) times of predicted moments for original and randomized videos. We observed that most SCDM's predictions are immune to the difference in input videos on ActivityNet. The distributions of differences of predictions can be found in supplementary material.
As for 2D-TAN, the performances for the original and randomized input videos are almost the same on both datasets.
Especially on ActivityNet Captions, 67.9\% of predictions remain exactly the same.

Figure~\ref{fig:performance_per_duration} shows the numbers of success and failure cases with respect to duration of ground-truth moments for the Action-Aware Blind baseline and 2D-TAN.
We can see that the performance gain of 2D-TAN mainly comes from long moments, but there is little differences between the baseline and 2D-TAN in shorter moments.
This result provides a possible explanation why 2D-TAN can achieve a good performance score even though it is less likely to make use of the visual information from input videos.
That is, if a model can guess that a given query sentence is likely to match a long moment, the model can easily obtain a high IoU by predicting a long moment, which can be the case for 2D-TAN as it gives high success rates for moments with longer duration. Together with 2D-TAN's immunity to input videos, this may means that, instead of finding semantic matches between the query sentence and input video, a deep model learn dataset biases and predict temporal locations of moments based only on query sentences. 

\begin{table}[t!]
\centering
\begin{tabular}{lcccc}
\toprule
Dataset          & Rep. Human & Random Human       & 2D-TAN & SCDM \\ \midrule
Charades-STA      & 52.1       & 42.8 (1.05) & 39.7   & 51.7 \\
ActivityNet Cap. & 44.4       & 35.4 (1.17) & 44.8   & 35.5 \\ \bottomrule
\end{tabular}
\caption{R@1(IoU$>$0.5) scores for 2D-TAN \cite{2DTAN_2020_AAAI} and 
SCDM \cite{yuan2019semantic}, as well as the human annotators. Rep.~Human and Random Human are the scores with (i) and (ii), respectively. Bracketed values for Random Human are the standard deviations over 100 trials.}
\label{tbl:human_performance}
\end{table}
\section{Human performance}
\label{sec:human_performance}
The experiments with our blind baselines suggest that recent models do not benefit much from the visual information. This raises a question: Are there any factors that prevent utilization of input videos. To answer this question, we examine how humans perform in the same task. Unfortunately, each query comes with a single annotation, thus, it is not feasible to assess each annotator's performance on the datasets. We are also interested in how well human annotators agree about moment boundaries. Severe disagreement may imply that answering correct moments are difficult even for humans possibly because there are some ambiguity in queries and videos. For deeper investigation of human performance, we re-collected temporal annotations with Amazon Mechanical Turk for Charades-STA and ActivityNet Captions.

We asked the annotators to work on the moment retrieval task,
where a query sentence and a video were displayed to an annotator, and the annotator marked the start and end times of a moment that corresponds to the query sentence. 
More details including the annotation interface can be found in the supplementary material.
We picked out random subsets from the test/validation sets and collected 5 answers for each sample (\ie, query-video pair).
We collected 5,000 annotations (1,000 samples) for Charades-STA and 6,440 annotations (1,288 samples) for ActivityNet Captions.
77 annotators were involved in total.

We assessed the human performance on the video moment retrieval task in two ways.
(i) We choose one representative annotator for each sample.
To obtain one representative annotator out of 5, we compute the pairwise IoUs among the moments from the 5 annotations (consequently, each annotator get 4 IoUs)  and select one annotator as representative that have the largest average IoU.
Representative annotators therefore are ones who are the most consistent with others.
(ii) We randomly sample 1 annotator out of 5 for each sample and compute R@1(IoU$>$0.5) for the entire test set. We compute the score for 100 times and report their average.

Interestingly, the results show that the scores of human annotators are lower or on a par with the state-of-the-art models (Table~\ref{tbl:human_performance}), even though it is fair to expect that our human annotators fully understand videos and query sentences. The limited human scores may suggest that the benchmark is not appropriately designed to assess the capability in video and language understanding. The dataset biases can partially explain why humans and deep models obtain similar scores: Deep models can exploit the priors learned from training samples, which increases models' scores. Meanwhile, humans do not have access to such prior and thus lower the scores, but they instead gain thanks to the visual information. 

\begin{figure}[t!]
    \centering
    \includegraphics[clip,width=\linewidth]{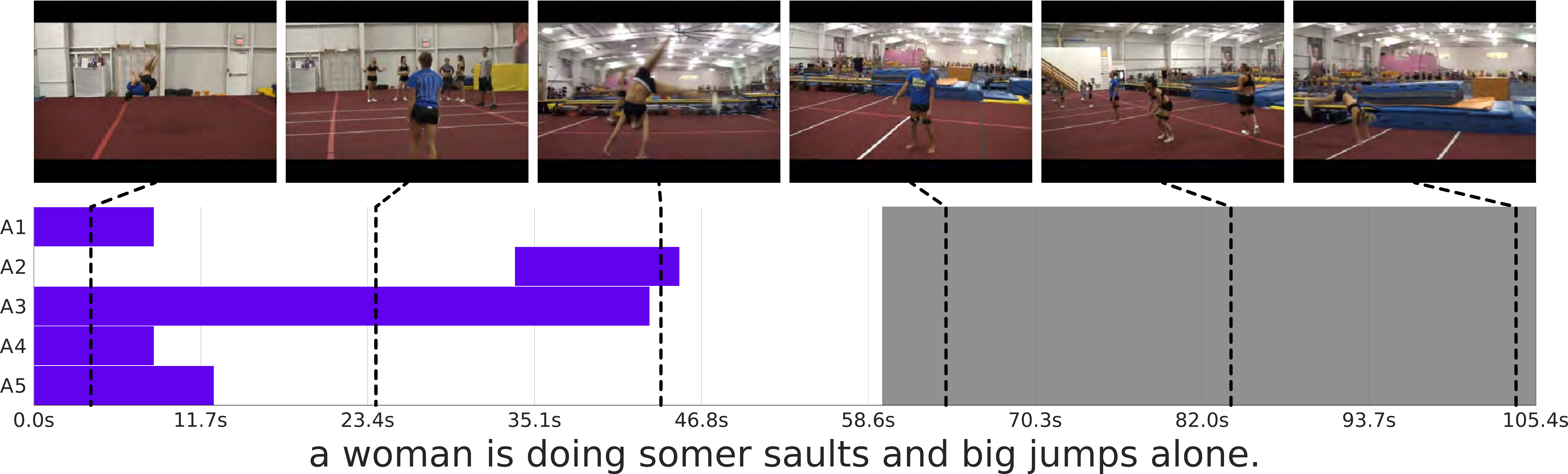}
  \caption{Example of temporal annotations on ActivityNet Captions. Blue bars indicate re-annotated moments, and gray area indicates ground truth moment.}
  \label{fig:human_annotation_examples}
\end{figure}

Another issue is in the current evaluation metric.
$\mathrm{R}@k(\mathrm{IoU}>m)$, which computes overlap between a predicted moment and a single ground-truth moment, ignores the ambiguity in the ground-truth moments.
That is, videos can have multiple moments that can be described by a query sentences. Ideally such moments should also be counted as positive matches, but the metric fails to do so.
Figure~\ref{fig:human_annotation_examples} shows an example in which a video has different positive moments.
The blue bars represent moments by new annotators, and the gray area represent the ground truth provided in ActivityNet Captions.
The video shows gymnasts practicing jumps repeatedly, and all re-collected moments match to the query sentence to some extent.

We also observed some unreliable ground-truth moments in the datasets.
For example, in Fig.~\ref{fig:human_annotation_examples}, the ground-truth moment actually shows ``three'' gymnasts performing jumps, while the query sentence says ``alone.''
Discarding such miss-labeled samples might be necessary for reliable evaluation.

\section{Alternative evaluation metrics}
Evaluation based on just a single ground-truth moment per sample can be problematic as (1) a video can have multiple positive moments and (2) there can be miss-labeled samples. To alleviate these problems, we present two alternative evaluation metrics that can exploit multiple annotated moments as references.

The first metric evaluates predicted moments with respect to the nearest-neighbor reference. This is based on the fact that a video may have multiple positive moments for a single query sentence.
When a predicted moment is close to at least one reference moment, the moment is counted as positive.
Specifically, a moment is counted as positive when the largest IoU among all reference moments exceed $m$.

The second metric considers the reliability of human annotations. When a reference moment largely overlaps with the majority of other reference moments, the reference moment is more reliable.
On the other hand, a reference moment that is different from others is possibly miss-labeled. Based on this assumption, we select the reference moment that has the largest average IoU to other moments, as representative reference.

$\mathrm{R}@1(\mathrm{IoU}>0.5)$ scores with the nearest neighbor and representative references are shown in Table~\ref{tab:multi_ref_evaluation}.
We evaluated Action-Aware Blind, 2D-TAN, SCDM, and Human annotations described in Sec.~\ref{sec:human_performance}.
GT represents ground truth references, either nearest neighbor (NN) or representative (Rep).
To evaluate human performance, we randomly select one annotation for each sample.
Note that when computing the score for human with the representative reference, we exclude the annotation used as representative and compute the score based on randomly selected one out of the remaining 4 annotations. The score is computed over the entire test set for 100 times, and their average and standard deviation are reported. 
The overall tendency of the performance remians the same for Action-Aware Blind, 2D-TAN, and SCDM with our metrics, but human annotators clearly outperform others.

\begin{table}[t!]
    \centering
\begin{tabular}{llcccc}
\toprule
Dataset              & GT   & \begin{tabular}[c]{@{}c@{}}Action-Aware\\ Blind\end{tabular} & \multicolumn{1}{c}{2D-TAN} & SCDM  & Human       \\ \midrule
Charades-STA         & NN   & 36.50                                                        & \multicolumn{1}{c}{51.70}   & 59.48  & 83.3 (0.96) \\
                     & Rep. & 18.70                                                        & 24.90                       & 34.50 & 66.8 (1.16) \\
ActivityNet Captions & NN   & 33.51                                                        & 60.24                        & 50.40  & 72.4 (1.07) \\
                     & Rep. & 15.97                                                        & 36.55                       & 31.17 & 52.3 (1.00) \\ \bottomrule
\end{tabular}
\caption{$\mathrm{R}@1(\mathrm{IoU}>0.5)$  with respect to the nearest neighbor reference (NN) and  representative (Rep.) references . The values in parentheses are the standard deviations. }
    \label{tab:multi_ref_evaluation}
\end{table}

\section{Conclusion}
Recent works have boosted the evaluation scores on the query-based moment retrieval benchmarks. However, it has not been assessed if these developments reflect the true progress in the task. Our experiments revealed that the high evaluation score \textit{per se} is not necessarily be the indisputable evidence that the model actually works well in practice. The datasets provided in major benchmarks include latent biases, and deep models are really good at making use of them. The evaluation metric, which has been extensively used in the literature, is not necessarily reliable, and we proposed alternatives, demonstrating better human performances.

Our results suggest that there are two directions for improvement.
One is to make better datasets by collecting diverse queries and finding samples where annotators agree on temporal locations. Another direction is data augmentation to diminish the effects of biases. For example, we may truncate or concatenate input videos to diversify temporal locations. 

\subsection*{Acknowledgements}
This work is partly supported by JSPS Kakenhi No.~18H03264 and the Academy of Finland projects 327910 and 324346.

\bibliography{egbib}

\begin{thebibliography}{20}
\providecommand{\natexlab}[1]{#1}
\providecommand{\url}[1]{\texttt{#1}}
\expandafter\ifx\csname urlstyle\endcsname\relax
  \providecommand{\doi}[1]{doi: #1}\else
  \providecommand{\doi}{doi: \begingroup \urlstyle{rm}\Url}\fi

\bibitem[Agrawal et~al.(2016)Agrawal, Batra, and
  Parikh]{agrawal-etal-2016-analyzing}
Aishwarya Agrawal, Dhruv Batra, and Devi Parikh.
\newblock Analyzing the behavior of visual question answering models.
\newblock In \emph{Empirical Methods in Natural Language Processing (EMNLP)},
  pages 1955--1960, 2016.

\bibitem[Alwassel et~al.(2018)Alwassel, {Caba Heilbron}, Escorcia, and
  Ghanem]{Alwassel2018}
Humam Alwassel, Fabian {Caba Heilbron}, Victor Escorcia, and Bernard Ghanem.
\newblock Diagnosing error in temporal action detectors.
\newblock In \emph{European Conference on Computer Vision (ECCV)}, pages
  256--272, 2018.

\bibitem[Chen and Jiang(2019)]{DBLP:conf/aaai/ChenJ19a}
Shaoxiang Chen and Yu{-}Gang Jiang.
\newblock Semantic proposal for activity localization in videos via sentence
  query.
\newblock In \emph{The {AAAI} Conference on Artificial Intelligence}, pages
  8199--8206, 2019.

\bibitem[Dawson et~al.(2018)Dawson, Zisserman, and Nell{\aa}ker]{Dawson2018}
Mitchell Dawson, Andrew Zisserman, and Christoffer Nell{\aa}ker.
\newblock From same photo: Cheating on visual kinship challenges.
\newblock In \emph{Asian Conference on Computer Vision (ACCV)}, pages 654--668,
  sep 2018.

\bibitem[Escorcia et~al.(2019)Escorcia, Soldan, Sivic, Ghanem, and
  Russell]{1907.12763}
Victor Escorcia, Mattia Soldan, Josef Sivic, Bernard Ghanem, and Bryan Russell.
\newblock Temporal localization of moments in video collections with natural
  language.
\newblock In \emph{CoRR arXiv:1907.12763}, 14 pages, 2019.

\bibitem[Gao et~al.(2017)Gao, Sun, Yang, and Nevatia]{Gao_2017_ICCV}
Jiyang Gao, Chen Sun, Zhenheng Yang, and Ram Nevatia.
\newblock {TALL}: Temporal activity localization via language query.
\newblock In \emph{International Conference on Computer Vision (ICCV)}, pages
  5267--5275, 2017.

\bibitem[Goyal et~al.(2017)Goyal, Khot, Summers-Stay, Batra, and
  Parikh]{balanced_vqa_v2}
Yash Goyal, Tejas Khot, Douglas Summers-Stay, Dhruv Batra, and Devi Parikh.
\newblock Making the {V} in {VQA} matter: Elevating the role of image
  understanding in {V}isual {Q}uestion {A}nswering.
\newblock In \emph{Conference on Computer Vision and Pattern Recognition
  (CVPR)}, pages 6904--6913, 2017.

\bibitem[Hahn et~al.(2019)Hahn, Kadav, Rehg, and Graf]{Hahn2019}
Meera Hahn, Asim Kadav, James~M. Rehg, and Hans~Peter Graf.
\newblock Tripping through time: Efficient localization of activities in
  videos.
\newblock In \emph{CoRR arXiv:1904.09936}, 14 pages, 2019.

\bibitem[Hendricks et~al.(2017)Hendricks, Wang, Shechtman, Sivic, Darrell, and
  Russell]{hendricks17iccv}
Lisa~Anne Hendricks, Oliver Wang, Eli Shechtman, Josef Sivic, Trevor Darrell,
  and Bryan Russell.
\newblock Localizing moments in video with natural language.
\newblock In \emph{International Conference on Computer Vision (ICCV)}, 2017.

\bibitem[Jiang et~al.(2019)Jiang, Huang, Yang, and
  Yuan]{10.1145/3323873.3325019}
Bin Jiang, Xin Huang, Chao Yang, and Junsong Yuan.
\newblock Cross-modal video moment retrieval with spatial and language-temporal
  attention.
\newblock In \emph{ACM International Conference on Multimedia Retrieval
  (ICMR)}, page 217–225, 2019.

\bibitem[Krishna et~al.(2017)Krishna, Hata, Ren, Fei-Fei, and
  Niebles]{krishna2017dense}
Ranjay Krishna, Kenji Hata, Frederic Ren, Li~Fei-Fei, and Juan~Carlos Niebles.
\newblock Dense-captioning events in videos.
\newblock In \emph{International Conference on Computer Vision (ICCV)}, pages
  706--715, 2017.

\bibitem[Liu et~al.(2018)Liu, Wang, Nie, He, Chen, and
  Chua]{10.1145/3209978.3210003}
Meng Liu, Xiang Wang, Liqiang Nie, Xiangnan He, Baoquan Chen, and Tat-Seng
  Chua.
\newblock Attentive moment retrieval in videos.
\newblock In \emph{ACM Conference on Research \& Development in Information
  Retrieval (SIGIR)}, page 15–24, 2018.

\bibitem[Regneri et~al.(2013)Regneri, Rohrbach, Wetzel, Thater, Schiele, and
  Pinkal]{regneri-etal-2013-grounding}
Michaela Regneri, Marcus Rohrbach, Dominikus Wetzel, Stefan Thater, Bernt
  Schiele, and Manfred Pinkal.
\newblock Grounding action descriptions in videos.
\newblock \emph{Transactions of the Association for Computational Linguistics},
  1:\penalty0 25--36, 2013.

\bibitem[Sigurdsson et~al.(2016)Sigurdsson, Varol, Wang, Farhadi, Laptev, and
  Gupta]{Sigurdsson2016}
Gunnar~A. Sigurdsson, G\"{u}l Varol, Xiaolong Wang, Ali Farhadi, Ivan Laptev,
  and Abhinav Gupta.
\newblock Hollywood in homes: Crowdsourcing data collection for activity
  understanding.
\newblock In \emph{European Conference on Computer Vision (ECCV)}, pages
  510--526, 2016.

\bibitem[Sigurdsson et~al.(2017)Sigurdsson, Russakovsky, and
  Gupta]{Sigurdsson2017}
Gunnar~A. Sigurdsson, Olga Russakovsky, and Abhinav Gupta.
\newblock What actions are needed for understanding human actions in videos?
\newblock In \emph{International Conference on Computer Vision (ICCV)}, pages
  2137--2146, 2017.

\bibitem[Wang et~al.(2019)Wang, Huang, and Wang]{Wang_2019_CVPR}
Weining Wang, Yan Huang, and Liang Wang.
\newblock Language-driven temporal activity localization: A semantic matching
  reinforcement learning model.
\newblock In \emph{IEEE Conference on Computer Vision and Pattern Recognition
  (CVPR)}, pages 334--343, 2019.

\bibitem[Xu et~al.(2019)Xu, He, Plummer, Sigal, Sclaroff, and
  Saenko]{DBLP:conf/aaai/Xu0PSSS19}
Huijuan Xu, Kun He, Bryan~A. Plummer, Leonid Sigal, Stan Sclaroff, and Kate
  Saenko.
\newblock Multilevel language and vision integration for text-to-clip
  retrieval.
\newblock In \emph{The {AAAI} Conference on Artificial Intelligence}, pages
  9062--9069, 2019.

\bibitem[Yuan et~al.(2019{\natexlab{a}})Yuan, Ma, Wang, Liu, and
  Zhu]{yuan2019semantic}
Yitian Yuan, Lin Ma, Jingwen Wang, Wei Liu, and Wenwu Zhu.
\newblock Semantic conditioned dynamic modulation for temporal sentence
  grounding in videos.
\newblock In \emph{Advances in Neural Information Processing Systems
  (NeurIPS)}, pages 534--544, 2019{\natexlab{a}}.

\bibitem[Yuan et~al.(2019{\natexlab{b}})Yuan, Mei, and
  Zhu]{DBLP:conf/aaai/YuanM019}
Yitian Yuan, Tao Mei, and Wenwu Zhu.
\newblock To find where you talk: Temporal sentence localization in video with
  attention based location regression.
\newblock In \emph{The {AAAI} Conference on Artificial Intelligence}, pages
  9159--9166, 2019{\natexlab{b}}.

\bibitem[Zhang et~al.(2020)Zhang, Peng, Fu, and Luo]{2DTAN_2020_AAAI}
Songyang Zhang, Houwen Peng, Jianlong Fu, and Jiebo Luo.
\newblock Learning {2D} temporal adjacent networks for moment localization with
  natural language.
\newblock In \emph{The {AAAI} Conference on Artificial Intelligence}, 2020.

\end{thebibliography}

\newpage
\renewcommand\thefigure{\Alph{figure}}    
\setcounter{figure}{0}    
\section*{Supplementary Material}
\section*{Data annotation}
\begin{figure}[h]
    \centering
    \includegraphics[clip,width=0.7\linewidth]{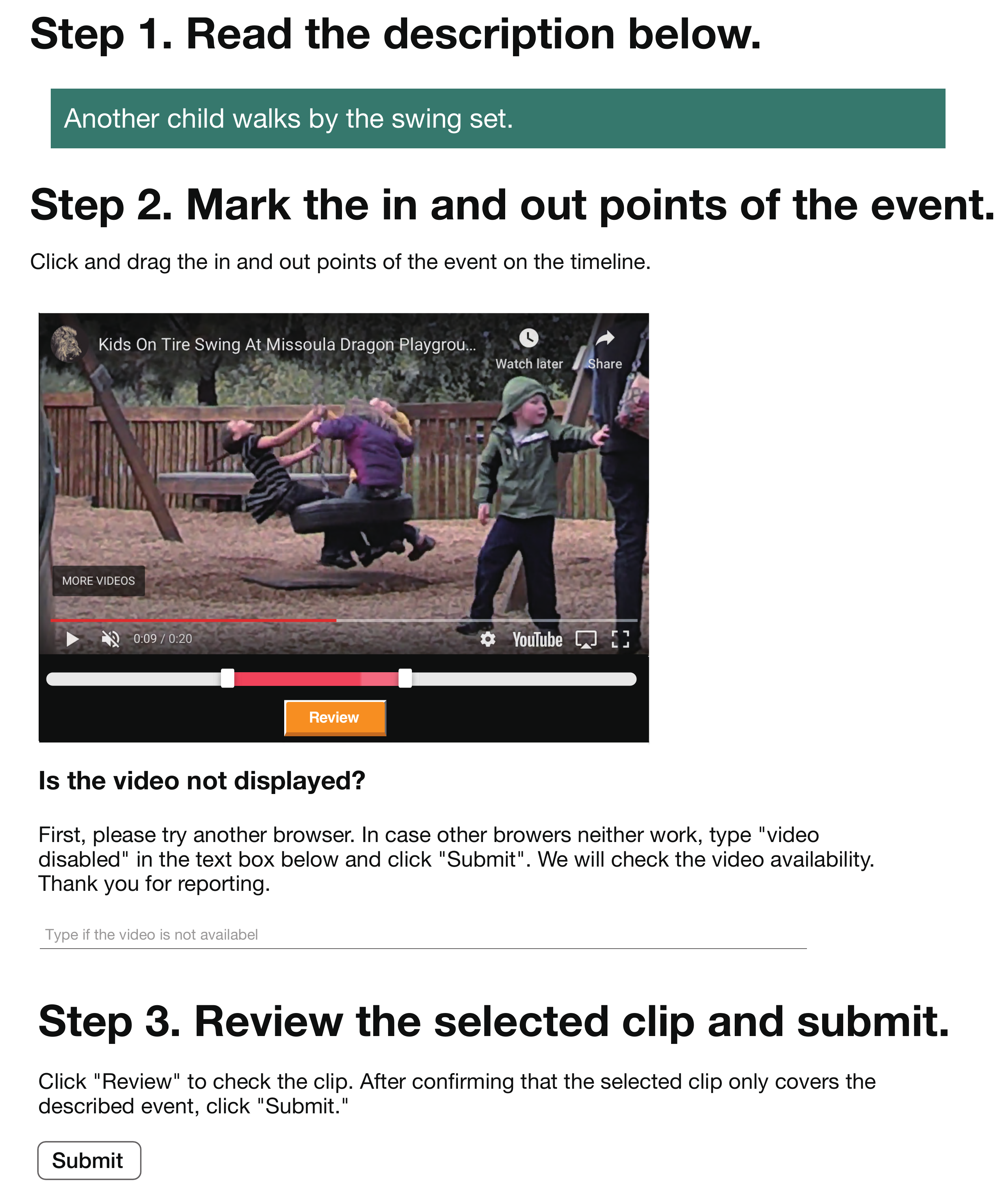}
    \caption{Annotation interface. Annotators drag the sliders on the control panel and mark the start and end points of moment. Clicking ``Review'' button plays the selected moment.}
    \label{fig:interface}
\end{figure}

\begin{figure}[t!]
    (a)\\ \includegraphics[clip,width=0.9\linewidth]{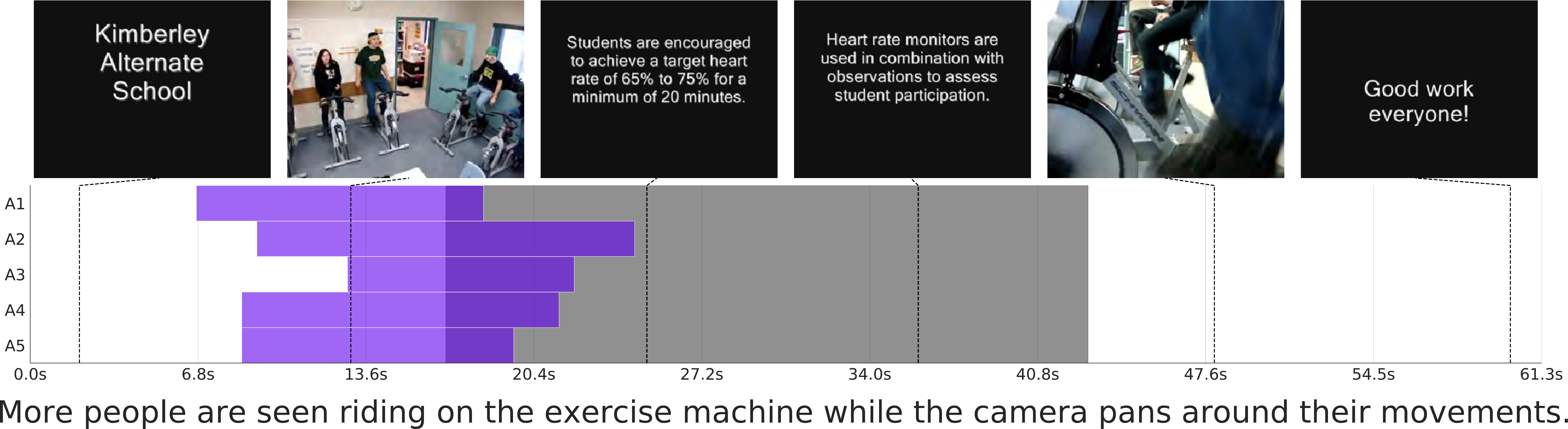}\\
    (b)\\ \includegraphics[clip,width=0.9\linewidth]{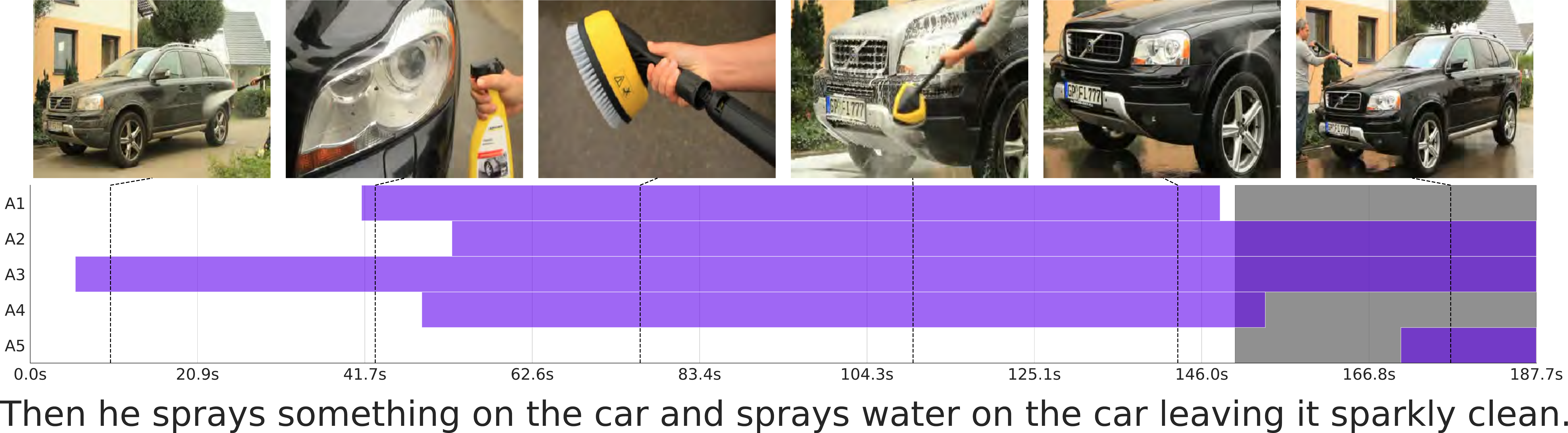}\\
    (c)\\ \includegraphics[clip,width=0.9\linewidth]{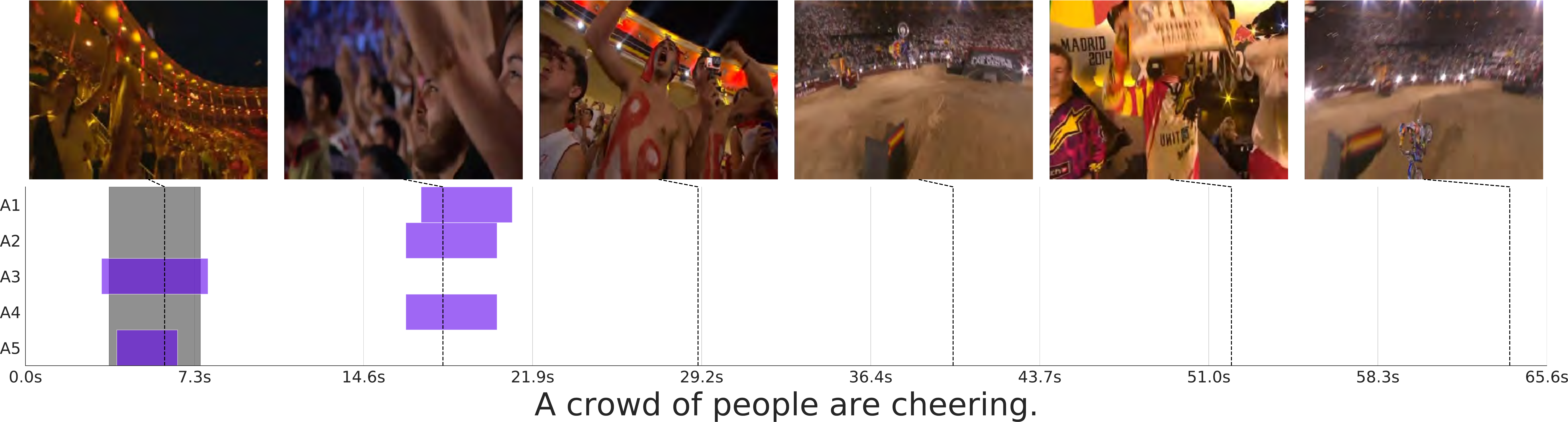}\\
    (d)\\ \includegraphics[clip,width=0.9\linewidth]{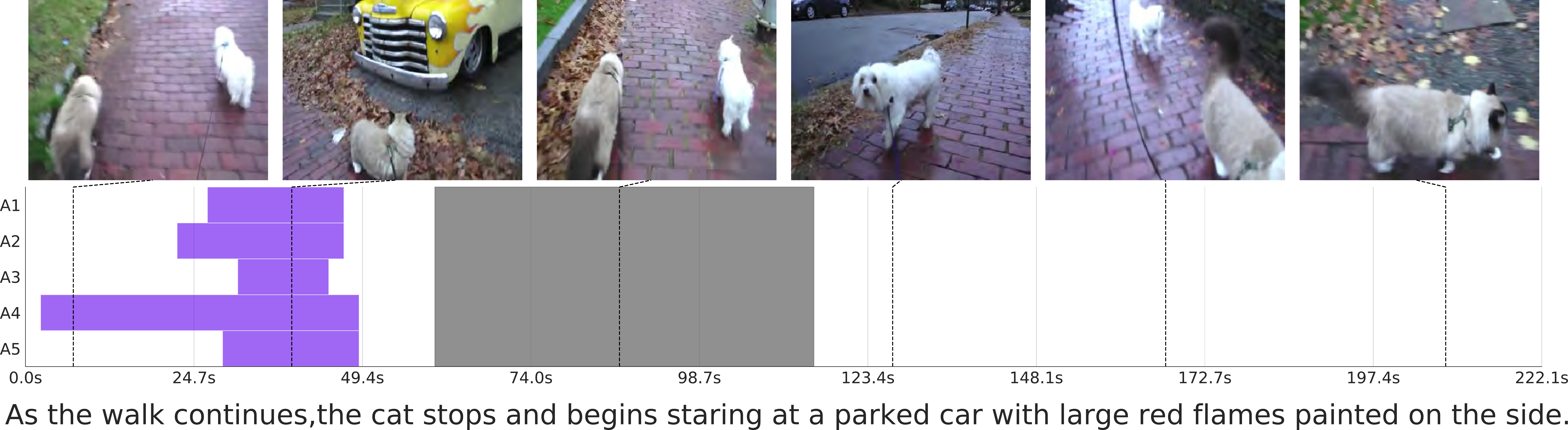}\\
    (e)\\ \includegraphics[clip,width=0.9\linewidth]{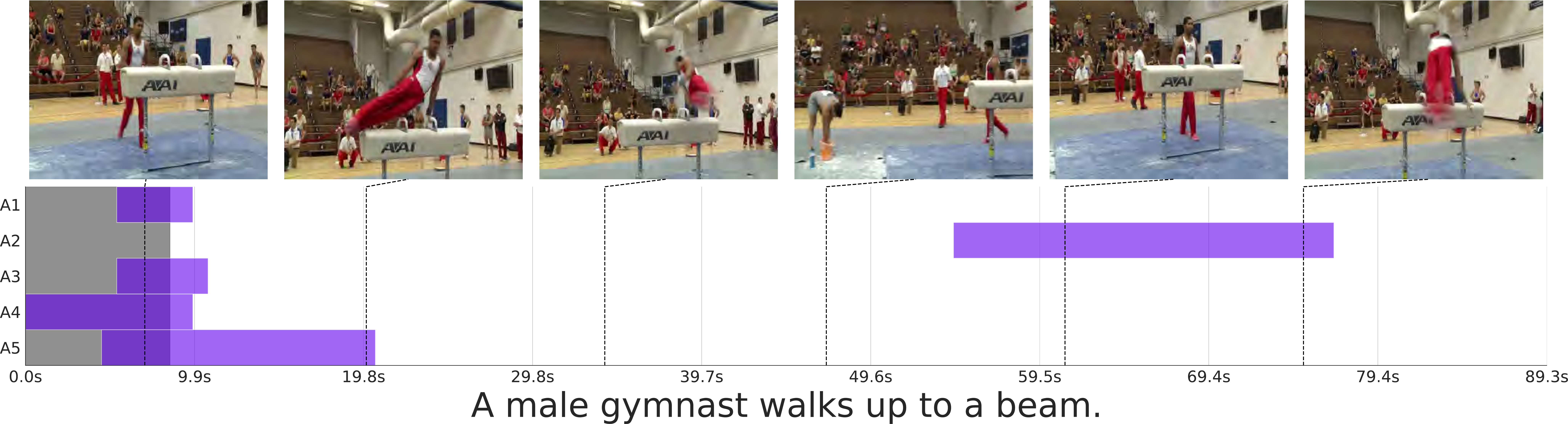}
    \caption{Examples of re-annotated moments on ActivityNet Captions.}
    \label{fig:activitynet_annotation_example}
\end{figure}

We recruited annotators on Amazon Mechanical Turk.
Figure~\ref{fig:interface} shows the annotation interface.
Average time to complete one annotation is about 8 minutes.
Some examples of annotated moments are in Fig~\ref{fig:activitynet_annotation_example}.
The purple bars represent re-annotated moments and gray area represents a ground truth moment provided by the original dataset.
We can see that the five annotators are likely to agree each other for some extent.
On the other hand, the re-annotated moments sometimes do not overlap with ground truth.
ActivityNet Captions dataset are created by writing a paragraph that describes the whole video and by annotating the temporal location in the video for each sentence.
This annotation procedure is different from actual video moment retrieval task, because the original annotators are exposed to the surrounding sentences in the paragraph, which contextualizes the sentence, whereas on only a single sentence is available for the model for prediction as well as for our annotators.

\section*{Distributions of temporal locations}
\begin{figure}[h!]
    \centering
    \includegraphics[clip,width=\linewidth]{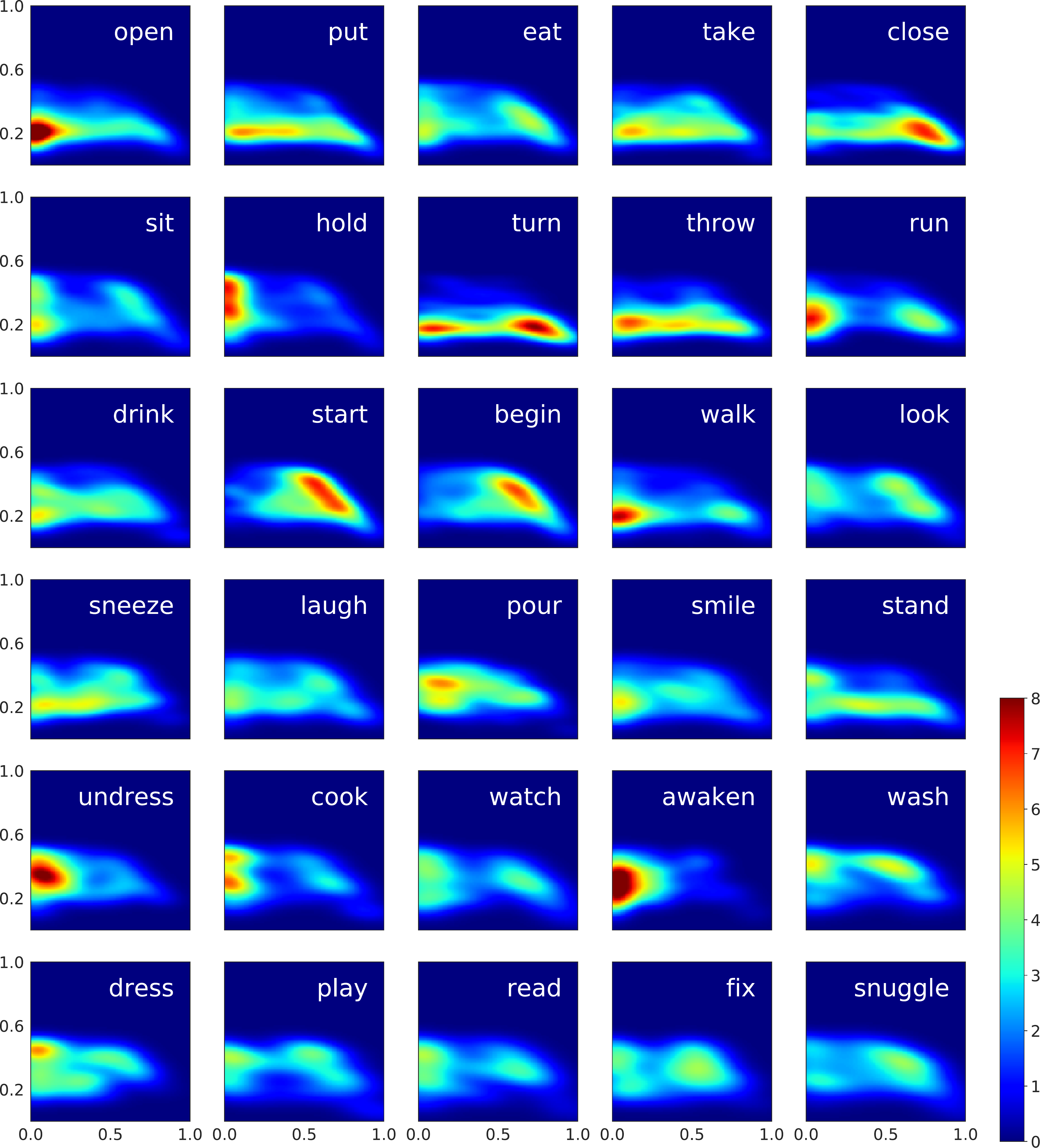}
    \caption{Distributions of temporal locations of target moments on Charades-STA. The distributions are generated for the verb on top right. Color represents values of probability density functions.}
    \label{fig:top30_priors_charade}
\end{figure}

\begin{figure}[h!]
    \centering
    \includegraphics[clip,width=\linewidth]{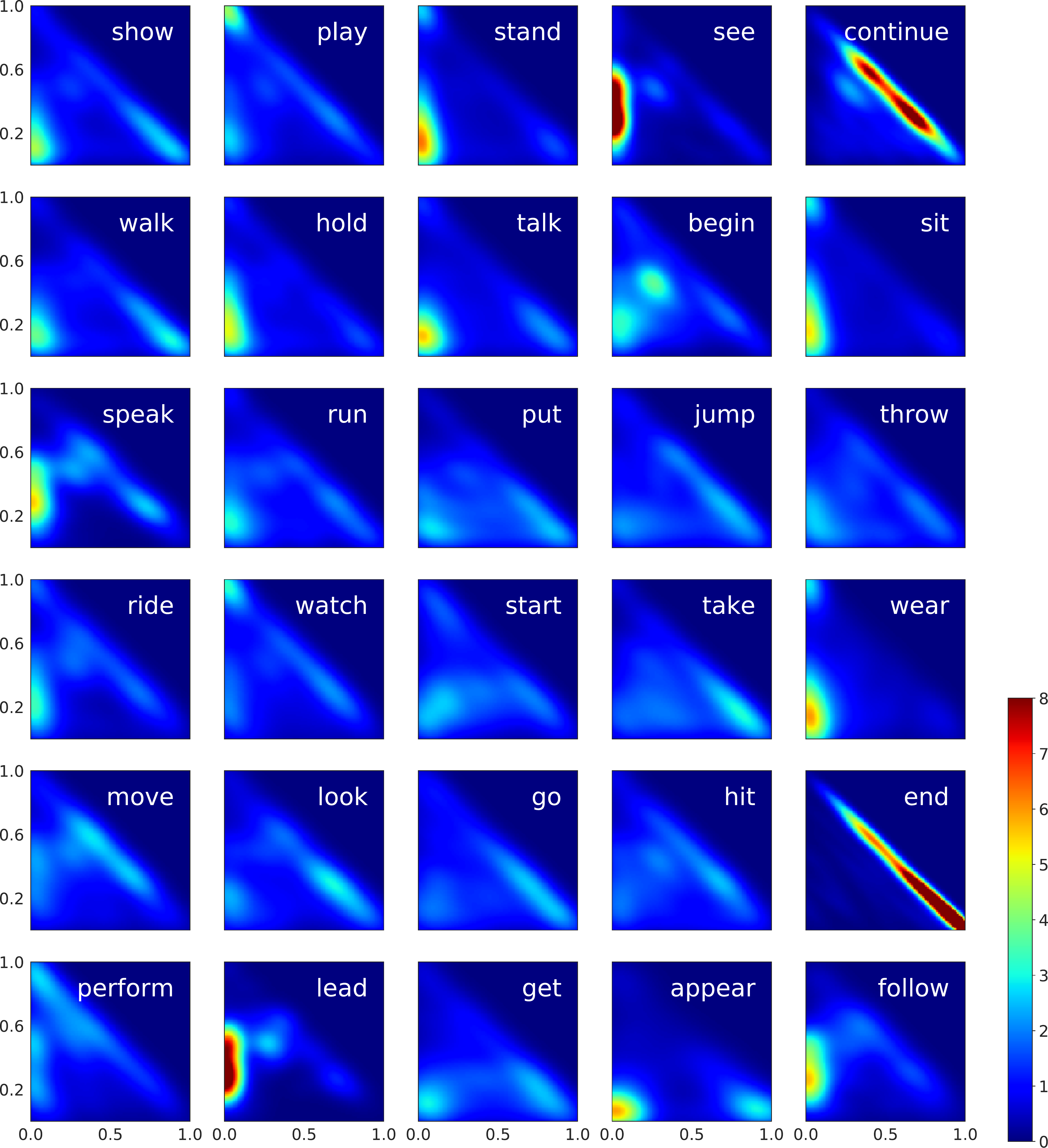}
    \caption{Distributions of temporal locations target moments on ActivityNet Captions. The distributions are generated for the verb on top right. Color represents values of probability density functions.}
    \label{fig:top30_priors_activitynet}
\end{figure}

Figure~\ref{fig:top30_priors_charade} and \ref{fig:top30_priors_activitynet} shows prior distributions of locations of moments. 
Each plot show a prior distribution of moments described by a verb on top right.
We selected the top-30 frequent verbs of each dataset.
Horizontal and vertical axes represent the start time and duration of a moment.

\section*{Blind-TAN}
\begin{figure}[t!]
    \centering
    \includegraphics[clip,width=\linewidth]{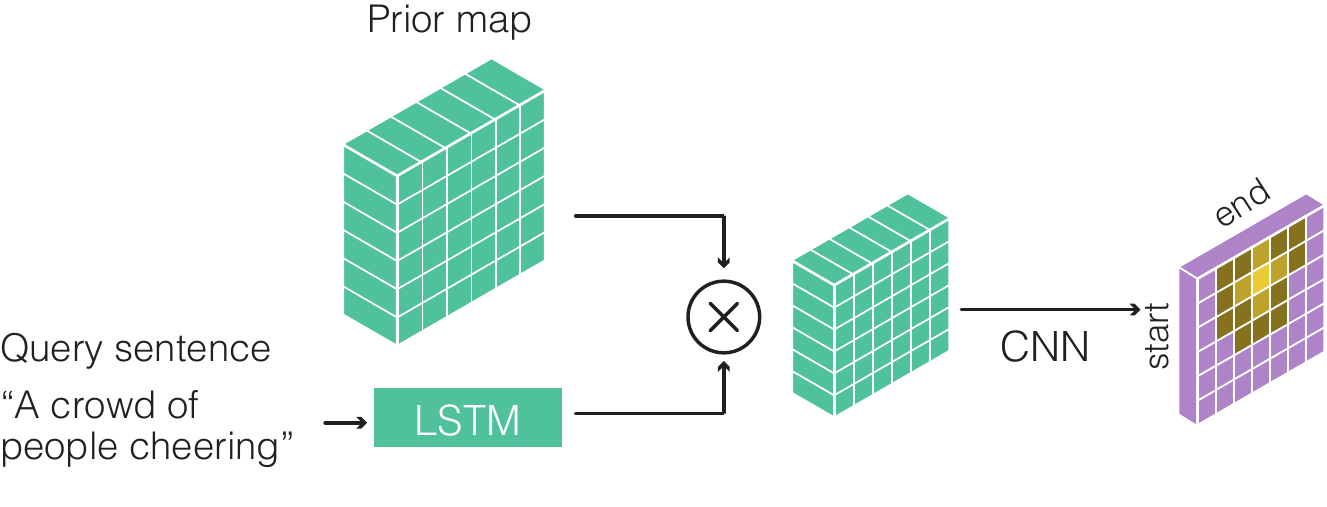}
    \caption{Overview of Blind-TAN, which is a reduced version of 2D-TAN \cite{2DTAN_2020_AAAI}. Blind-TAN only takes a query sentence as input and predicts where the target moment is likely to locate.}
    \label{fig:blind-tan}
\end{figure}{}

To build Blind-TAN, we remove the component to handle input video in 2D-TAN, and instead, add a map $M_p\in\mathbb{R}^{N \times N \times d}$ where $N$ is the number of sampling points of start and end times, and $d$ is the channel size.
We set $N$ to 256 for both Charades-STA and ActivityNet Captions, and  $d$ to 512 for Charades-STA and 128 for ActivityNet Captions.
Note that the prior map is learnable.
Query sentence feature is compute by GloVe word embedding and LSTM layers.
The query sentence feature and the prior map is fused by the Hadamard product.
A multi-layer CNN transform the fused map into 2D score map. 
The score of a moment starting from point $i$ and ent at point $j$ is represented by the score map value at $(i, j)$.

Blind-TAN is trained on pairs of a query sentence and a ground-truth moment location.
Videos are not used for training nor testing.
The prior map is initialized with random values and updated by backpropagating errors between the predicted score map and ground-truth.
The loss function is a weighted binary cross-entropy as in \cite{2DTAN_2020_AAAI}.

\section*{The effects of randomized videos}
\begin{figure}[t!]
    \centering
    \includegraphics[clip,width=0.7\linewidth]{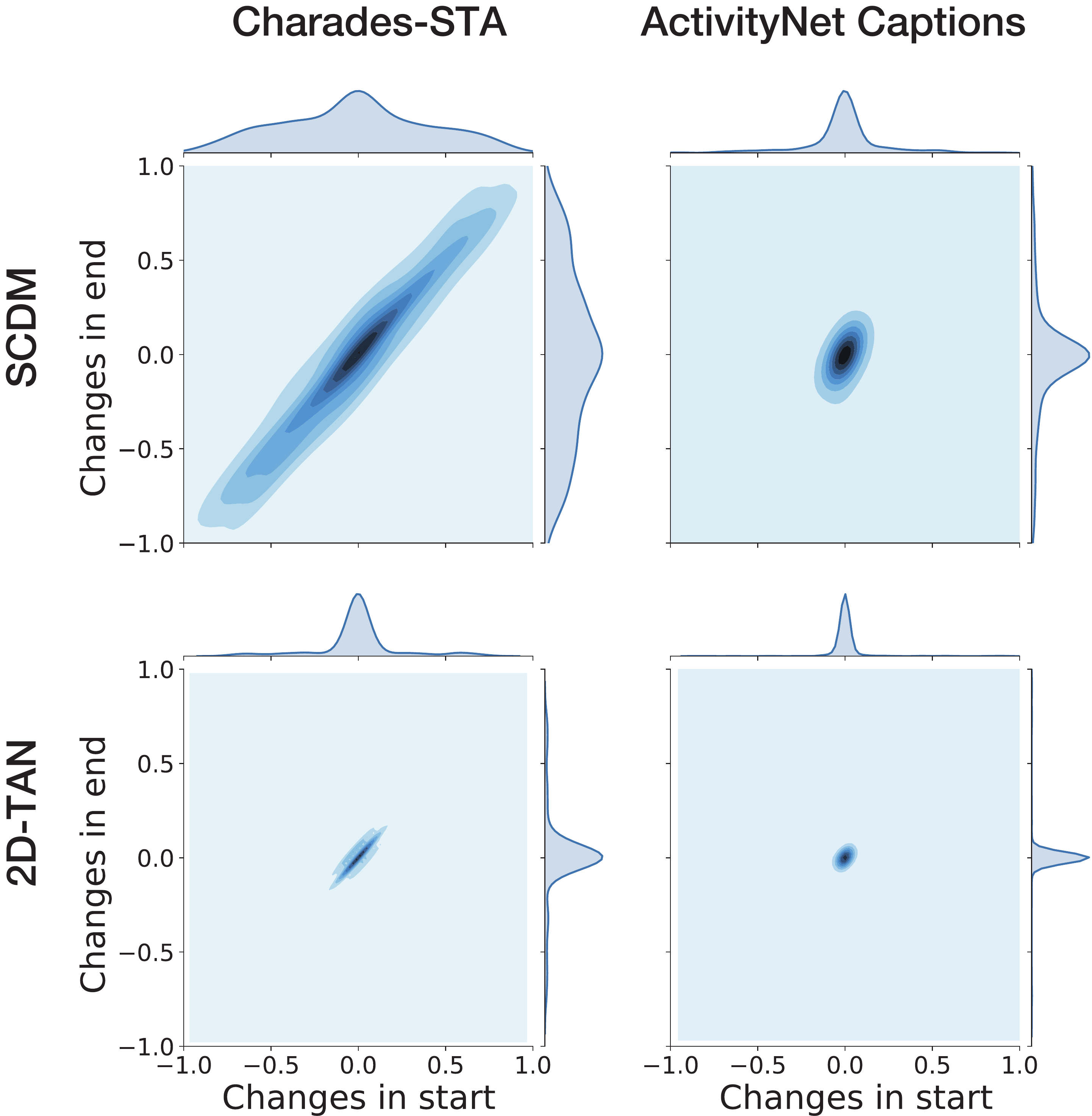}
    \caption{Differences between temporal locations of moment predicted using original videos and randomized videos.}
    \label{fig:output_change}
\end{figure}

We provide detailed results on how randomized videos change output moments.
On the randomization test in Sec.~\ref{sec:sanity_check}, the deep models predict moment boundaries using original videos and randomized videos.
Let $(s_i, e_i)$ be start and end points of original prediction, and $(s'_i, e'_i)$ be those on randomized video, where all values are normalized to the range $[0,1]$ by dividing them by the length of the respective video.
We compute how randomization affect the prediction by computing $d^s_i = |s_i - s'_i|$ and $d^e_i = |e_i - e'_i|$.
Figure~\ref{fig:output_change} shows the joint distributions of the differences $d^s_i$ and $d^e_i$, where the Gaussian kernel density estimation is used to generate the distributions; the horizontal and vertical axes are $d^s_i$ and $d^e_i$, respectively.
Except for SCDM on Charades-STA, the models hardly change the output for randomized videos.
This demonstrates that 2D-TAN and SCDM trained on ActivityNet Captions actually ignore input videos.
Only SCDM on Charades-STA is affected by randomization.
However, the start and end times change with similar amounts. That is, the predicted duration of moments are likely to be constant regardless of input videos.
This tendency suggests that SCDM guesses the duration of target moments only using priors.

\end{document}